\documentclass[a4paper,conference]{IEEEtran}

\usepackage{graphics} % for pdf, bitmapped graphics files
\usepackage{amsmath} % assumes amsmath package installed
\usepackage{amssymb} % assumes amsmath package installed
\usepackage{tikzpagenodes}

\usepackage{url}

\usepackage[dvips]{rotating} 
\usepackage{array}
\usepackage{multirow}
\usepackage{float}
\usepackage{rotating}
\usepackage{subfigure}
\usepackage{hhline}
\usepackage{wrapfig}
\usepackage{dblfloatfix} % To enable figures at the bottom of page
\newcommand*\rot{\rotatebox{90}}
\begin{document}

\title{Two-stage CNN-based wood log recognition}

\author{\IEEEauthorblockN{Georg Wimmer and Rudolf Schraml and Heinz Hofbauer and Andreas Uhl}
\IEEEauthorblockA{University of Salzburg\\
Jakob Haringer Str. 2, 5020 Salzburg\\
Salzburg, Austria\\
Email: \{gwimmer, rschraml, hhofbaue, uhl\}@cs.sbg.ac.at}
\and
\IEEEauthorblockN{Alexander Petutschnigg}
\IEEEauthorblockA{University of Applied Sciences Salzburg\\
 Markt 136a, 5431 Kuchl, Austria}}

\IEEEoverridecommandlockouts
\IEEEpubid{\makebox[\columnwidth]{978-1-5386-5541-2/18/\$31.00~\copyright2018 IEEE \hfill} \hspace{\columnsep}\makebox[\columnwidth]{ }}

\maketitle

\IEEEpubidadjcol

% As a general rule, do not put math, special symbols or citations
% in the abstract
\begin{abstract}
The proof of origin of logs is becoming increasingly important. In the context of Industry 4.0 and to combat illegal logging there is an increasing motivation to track each individual log. Our previous works in this field focused on log tracking using digital log end images based on methods inspired by fingerprint and iris-recognition. This work presents a convolutional neural network (CNN) based approach which 
comprises a CNN-based segmentation of the log end combined with a final CNN-based recognition of the segmented log end using the triplet loss function for CNN training.
Results show that the proposed two-stage CNN-based approach outperforms traditional approaches. %inspired by fingerprint and iris-recognition.
\end{abstract}
% 
% \begin{keywords}
% \vspace{-1mm}
% log tracing, deep learning, segmentation
% \end{keywords}
%

%\IEEEpeerreviewmaketitle
%\footnote{Blabla}

\section{Introduction}

% A weng einleitung mit aktuellem Bezug
Methods for the tracing of logs are an essential component in solving a wide variety of problems and requirements of an ecological, legal and social nature. Currently, this mainly relates to proof the origin of wood products, e.g. by certification companies like the Forest Stewardship Council (FSC). However, efforts towards traceability down to the individual tree log have been intensified by a variety of stakeholders. The motivation for this is that, on the one hand, illegal logging can be better combated and, on the other hand, the identification of each individual wood log forms a basis for steps towards forest-based Industry 4.0.
% State of the art 
In the context of Industry 4.0, Radio Frequency Identification (RFID) is the state-of-the-art for object recognition/tracking. However, like a set of other tracking technologies for wood logs (e.g. punching, coloring or barcoding log ends \cite{Tzoulis2013}), RFID requires physical marking of each tree which suffers costs.
% Unsere Arbeiten...
%
An alternative to physical marking is to use biometric characteristics to recognize each individual log. A short summary on biometric log tracking using various characteristics is presented in \cite{Schraml15d}. In a series of works between 2014--2016 we investigated wood log tracking based on digital log end images in regard to the distinctiveness and robustness of the annual ring pattern. For a literature review we refer to \cite{Schraml16a}.
%Figure \ref{fig:log_traceability} presents the scheme of such a log tracking system.
%Figure ??? oder nicht
%\begin{figure}[bt]
%	\centering
%	\includegraphics[width=0.98\columnwidth]{mva_system}
%	\caption{Exemplary scheme of enrolment and identification for wood log tracking}
%	\label{fig:log_traceability}
%\end{figure}
Significant for this work is that the utilized approaches were inspired by human fingerprint and iris-recognition methods. Those rely on traditional feature extraction methods (e.g. Gabor filterbanks) and moreover require a sophisticated pre-processing (segmentation, pith estimation, rotational pre-alignment) of each log end image prior to feature extraction. Comparison of the extracted features is also complex. Furthermore, it has to be noted that for our previous works manually segmented log end images and determined pith positions were utilized. 
%
% Aktuelle Referenzen DL im Bereich Holz/bildverarbeitung
%
Time has passed and deep learning based approaches have become state-of-the-art. Not surprisingly, deep learning-based methods have also been investigated in many application areas in the forestry and timber industry in recent years. Exemplary applications are wood species identification using cross-section (CS) images \cite{Tang2018,Olschofsky2020}, remote sensing-based tree species classification \cite{Fricker2019} or lumber grading \cite{Hu2019} using wood board surface images. 

In this work we apply convolutional neural networks (CNNs) for two-stage wood log recognition. CNNs are used for segmentation of the CS in the log end image as well as for feature extraction that offers advantages in many ways. The experimental evaluation is based on a database (DB) which was utilized in \cite{Schraml16c} and a new DB denoted as 100 logs DB (HLDB). This work significantly contributes to biometric log end recognition by showing that a CNN-based approach does not require to determine the pith position and moreover no rotational pre-alignment is required. Results show that CNN-based segmentation and feature extraction shows a similar performance as the results presented in \cite{Schraml16c} which are based on groundtruth data and traditional feature extraction methods. The experimental evaluation on the new HLDB, for which no groundtruth data is available, underpins this statement and shows the weaknesses of the traditional methods that are mainly caused by inaccurate segmentation and pith estimation results.
 \begin{figure*}[t!]
 	
 	\centering
 		\subfigure[Forest site - HDLB$_{FH,FL}$ \label{fig:forest}]{	\includegraphics[height=2.8cm]{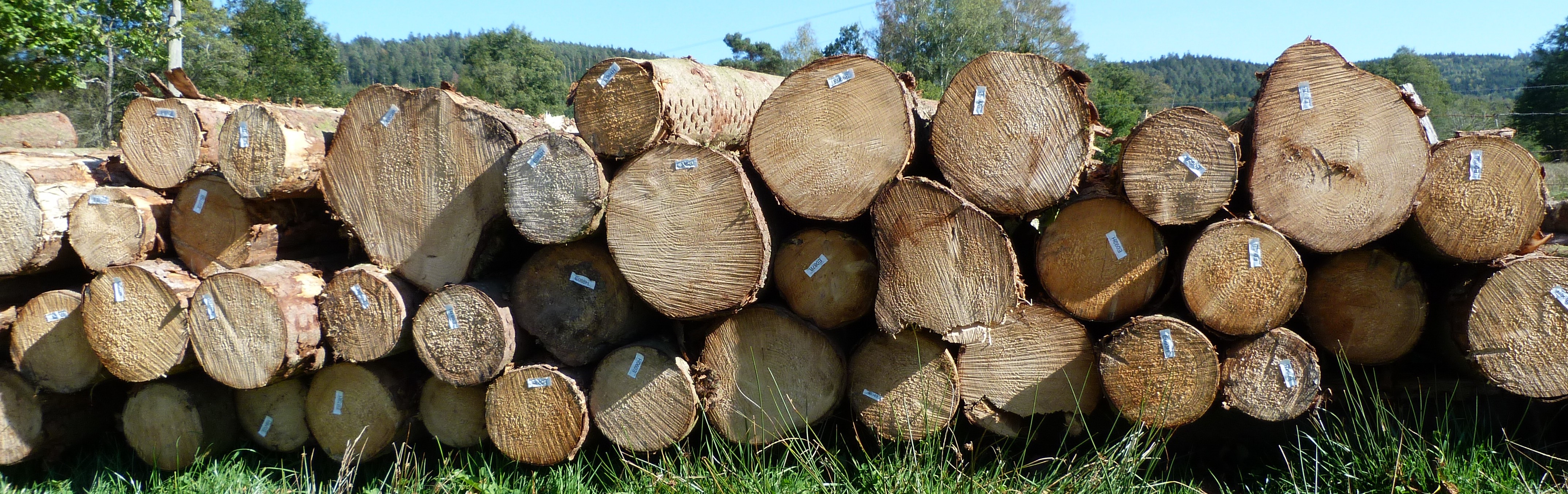}}
  \subfigure[Sawmill yard - HDLB$_{SM}$ - discs were cut for HLDB$_{R,S}$\label{fig:sawmill}]{	\includegraphics[height=2.8cm]{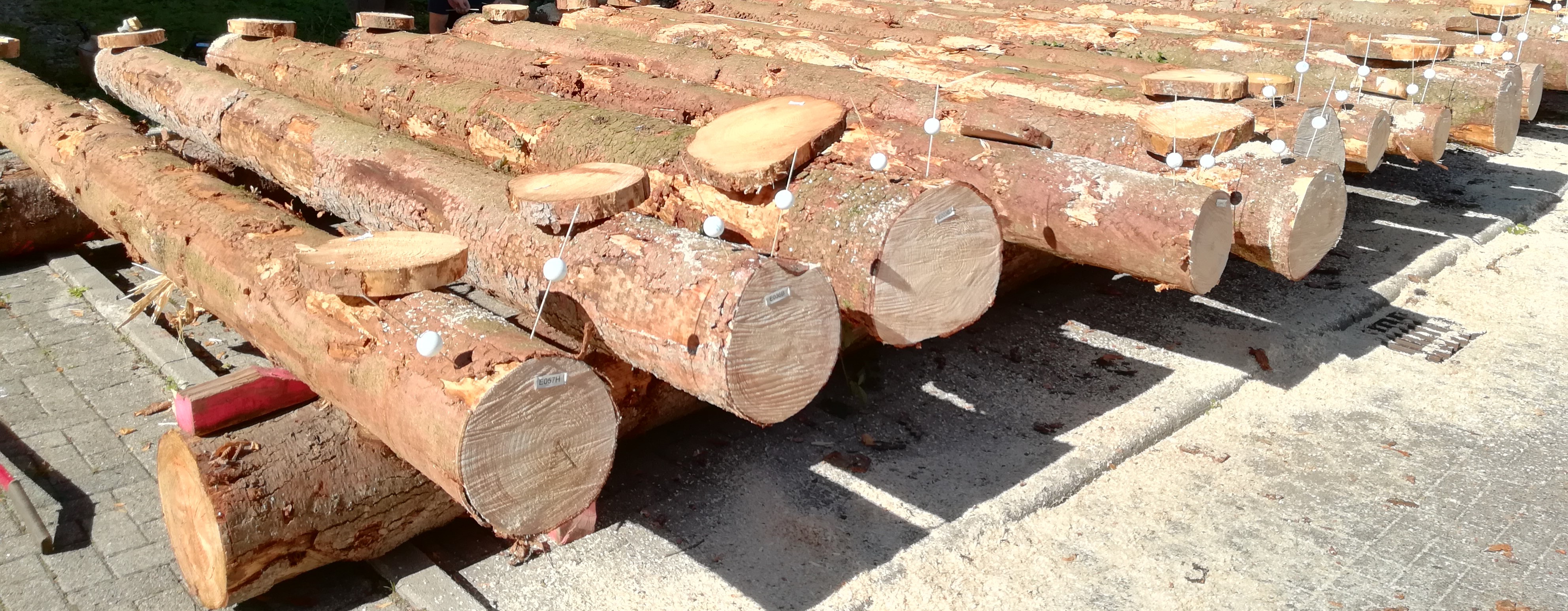}}

 \caption{
 %\vspace{-0.2cm}
 100 Logs Image Database (HLDB): Fig.~\ref{fig:forest} shows log piles close to the forest at which the forest datatsets were acquired. Fig.~\ref{fig:sawmill} shows the data acquisition at the sawmill yard where discs of each log end were cut off.}
 \label{fig:hldb}

 \end{figure*}

%\begin{figure}[t!]
% 	\centering
% 		\subfigure[Forest site - HDLB$_{FH,FL}$ \label{fig:forest}]{	%\includegraphics[width=0.47\columnwidth]{illustration/100logs_forest.jpg}}
% \subfigure[Sawmill yard - HDLB$_{SM}$\label{fig:sawmill}]{	\includegraphics[width=0.47\columnwidth]{illustration/100logs_sawmill.jpg}}
% \caption{
% %\vspace{-0.2cm}
% 100 Logs Image Database (HLDB): For 100 logs, %harvested in France, 
% different log end image datasets were acquired. Fig.~\ref{fig:forest} shows log piles close to the forest at which the forest datatsets were acquired. Fig.~\ref{fig:sawmill} shows the data acquisition at the sawmill yard where discs of each log end were cut off for HLDB$_{R,S}$.}
%\label{fig:hldb}
%\vspace{-0.5cm}
%\end{figure}

%
\def\singlew{0.19}
\begin{figure*}[b!]
	
	\centering
	\subfigure[HLDB$_{FH}$]{\includegraphics[width=0.17\textwidth]{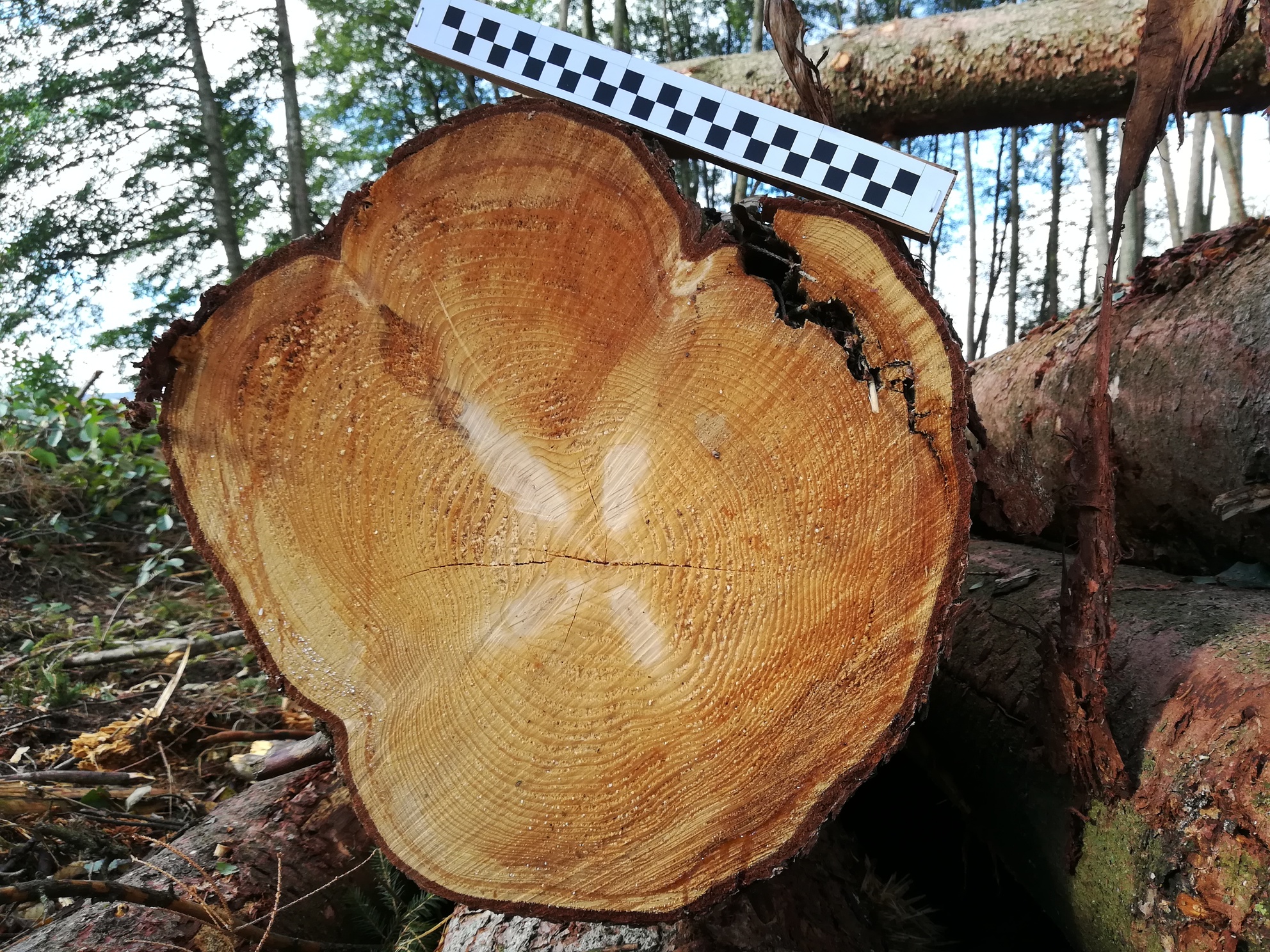}}
	\subfigure[HLDB$_{FL}$]{\includegraphics[width=0.17\textwidth]{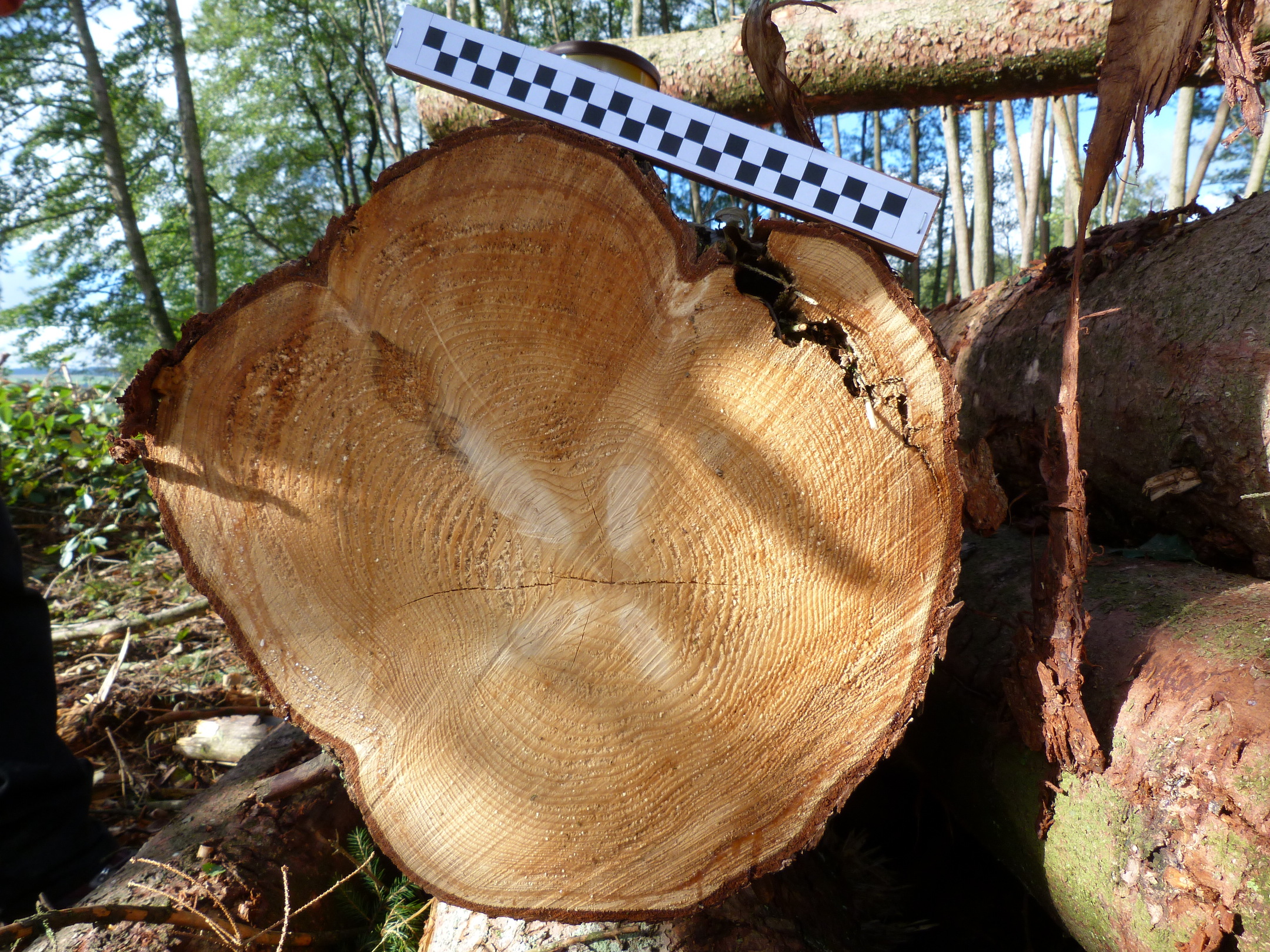}}
	\subfigure[HLDB$_{SM}$]{\includegraphics[width=0.17\textwidth]{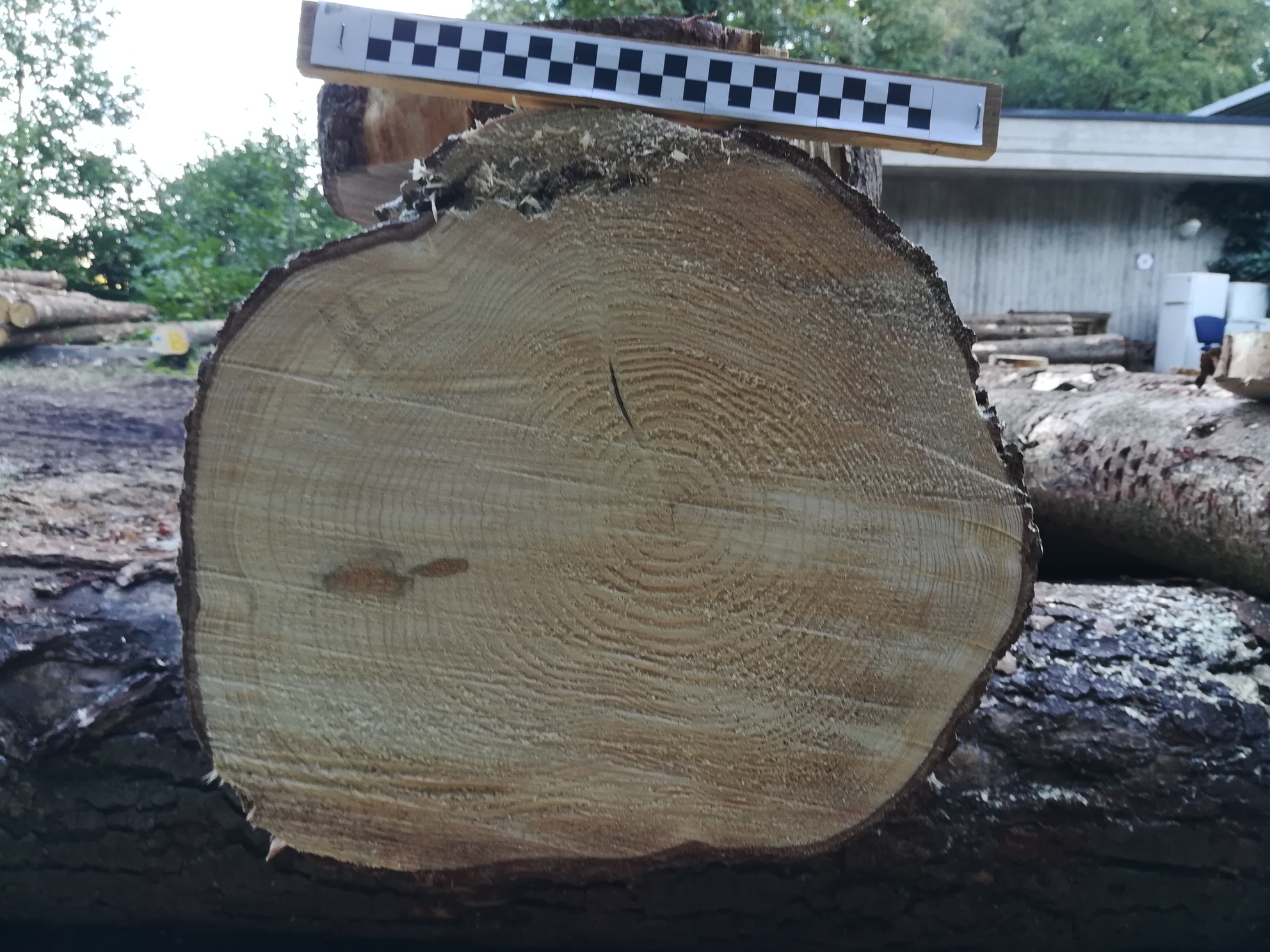}}
	\subfigure[HLDB$_{R}$]{\includegraphics[width=\singlew\textwidth]{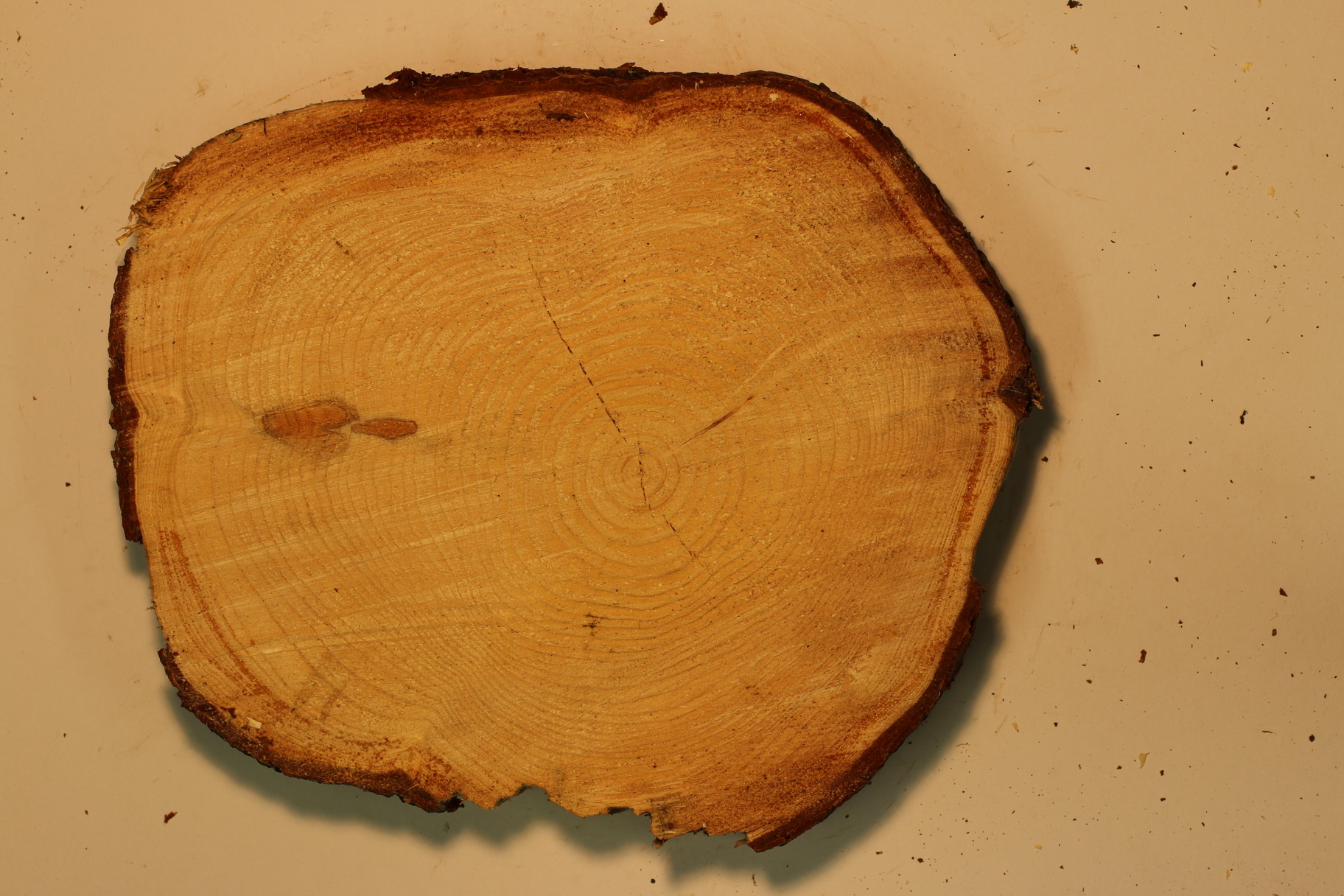}}
	\subfigure[HLDB$_{S}$]{\includegraphics[width=\singlew\textwidth]{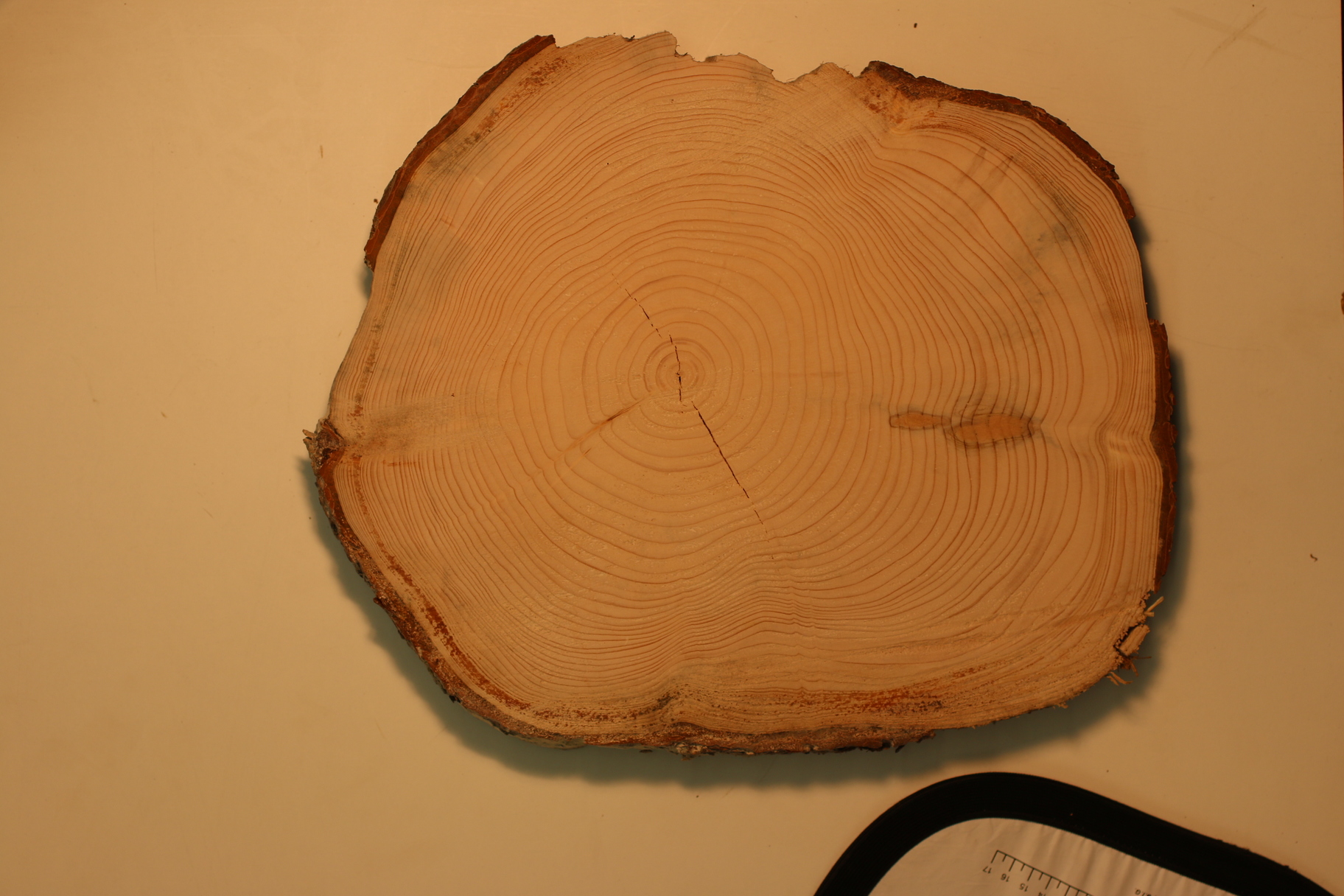}}
	\caption{HLDB: Exemplary images for both log ends of log \#E001 from all datasets HLDB$_{FH,FL,SM,R,S}$}
	
	\label{img:databaseSamples}
\end{figure*}

Sect.~\ref{sec:material} introduces the DBs. The CNN-based segmentation of the CS is explained in Sect.~\ref{sec:segmentation} followed by the CNN-based log recognition in Sect. \ref{sec:recognition}. Sect.~\ref{sec:setup} presents the experimental setup and Sect.~\ref{sec:results} the results.
Sect.~\ref{sec:conclusions} concludes this work.

\section{Material}
\label{sec:material}

Two different DBs are utilized: (i) the MVA DB which was already utilized in \cite{Schraml15c,Schraml16c} and (ii) a new database referred to as 100 Logs Database (HLDB). For a detailed description of MVA we refer to \cite{Schraml16c}. The MVA DB is utilized (i) to compare the CNN-based results to previous results which were based on annotated groundtruth data and (ii) to train a segmentation CNN in order to segment the images of HLDB, for which no manual segmentation is available. 
HLDB comprises different datasets which were all taken from the same 100 logs. CS-Images were acquired from both ends of each logs. The first two datasets HLDB$_{FH}$ and HLDB$_{FL}$ were taken in the forest (see Fig.~\ref{fig:hldb}) using a Lumix camera and a Huawei smartphone, respectively. Both datasets consist of 4 images for each log end. After two images the camera was rotated by approximately 45 degrees and two more images were captured. The next dataset, denoted as Sawmill dataset (HLDB$_{SM}$), was captured after cutting off a thin disc from each log end (see Fig. \ref{fig:hldb}). Three images with different rotations for each fresh cross-cut log end were acquired using the Huawei smartphone.
%which results in varying saw cutting patterns, possibly different branches in the cross section and a slightly different shape of the log cross section as for the two Forest datasets). 
The CS-Images of HLDB$_{FH,FL,SM}$ were taken without tripod which causes different rotations, slightly varying perspectives toward the CS and slightly different positions of the CS in the image. Finally, one side of the 200 discs was acquired using a Canon EOS 70D with a tripod and lighting, once raw (HLDB$_R$) and once after they were sanded (HLDB$_{S}$). The captured CSs are mirrored versions of the CSs in the Sawmill dataset. For HLDB$_R$ four and for HLDB$_S$ six CS-Images with different rotations were captured, respectively.
%CT scanners are used in big saw mills to optimize the sawmill processing. The logs are sent through the CT scanner and about all 4.5 mm along the length of a log a CT image is taken from the log CS. 
%zwischen 874 und 905 scans per log (durchschnittlich 889) auf wahrscheinlich gute 4m Länge.
%From all the images per log (889 image per log in average) we only employ the first and the last 15 images, these that are most close to one of the ends of the log. In that way this dataset is acquired in a similar way than the other datasets, where images were taken from the log ends. The images of a log all have the same rotation, scale and perspective, but they possibly show different branches in the images since the CT images were taken at different longitudinal positions on the log.
Figure \ref{img:databaseSamples} shows exemplary images for all datasets from the bottom end of log labelled \#E001. It can be observed that the CS-Images of the two forest datasets HLDB$_{FH,FL}$ look quite similar since the images were taken with the same surrounding, the same log cut pattern and hardly any time shift between the taking of the images of the two datasets. The CS-Images captured at the sawmill yard HLDB$_{SM}$ look completely different because of the fresh cut that results in a totally different saw cut pattern and wood coloration. The disc CS-Images HLDB$_{R,S}$ are captured under idealistic conditions and serve as a reference in the experiments, 
especially %when considering 
the sanded CS-Images in HLDB$_{S}$ which show a undisturbed annual ring pattern.

\section{CS-Segmentation}

\label{sec:segmentation}
Prior to any feature extraction the CS area in the CS-Image needs to be localized and segmented from the background. We apply the Mask R-CNN framework \cite{He2017} to get a segmentation mask.
%For the segmentation of the log CS, we apply the Mask R-CNN framework \cite{He17} to segment the log CS by predicting the segmentation mask. % and recognize the bounding box. 
As net architecture we employ the ResNet-50 architecture using a model pretrained on the COCO dataset. The segmentation net is then trained on MVA for which segmentation mask groundtruth is available (Sect.~ \ref{sec:setup}). The segmentation net is trained for 30 epochs in order to differentiate the CS from the background.
Then, the trained segmentation net is applied to the HLDB datasets to segment the CS from the background. To also get CNN-based segmentation masks for MVA, we apply a 4-fold cross validation, where one fold consists of a fourth of the 279 logs (the images of one log are all in the same fold). 3 folds are used to train the segmentation net and the trained net is applied to segment the images of the remaining fold. 
%The segmentation of the CT images is not possible using a segmentation net, since we have no manually segmented CT images to train a CNN and the RGB-images of the MVA database are too different to the CT images. %to be able to segment the CT images using a net trained on MVA images. 
%However, the CT images already have a black blackgound outside of the log CS (see Figure \ref{img:databaseSamples}(g,h)) and so segmentation is easy and has been applied using the active contour method \cite{Chan01}.
%
\begin{figure}[b!]

	\centering
	\includegraphics[width=1.0\columnwidth]{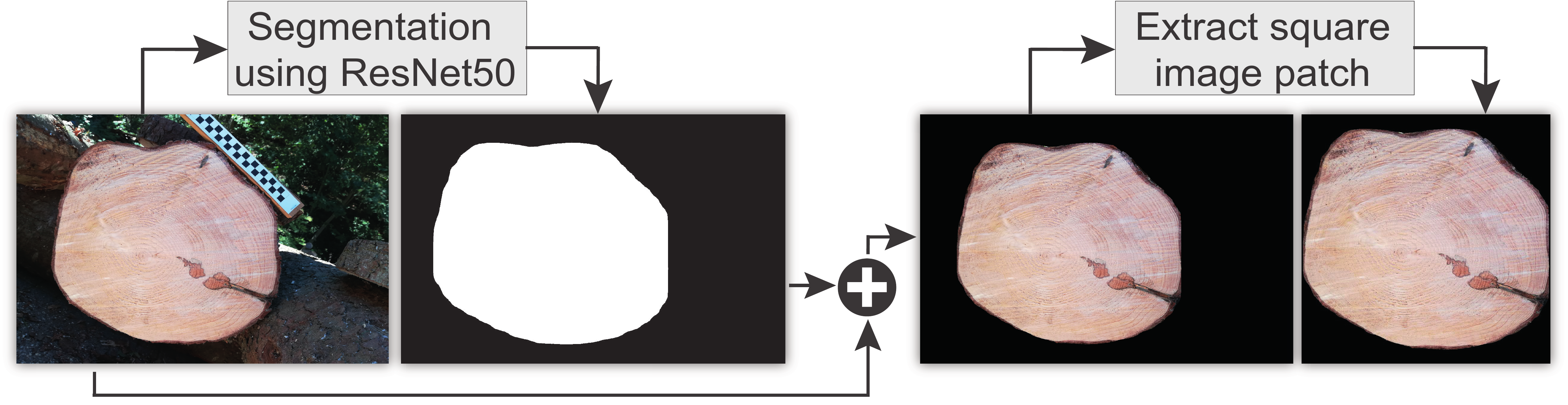}
	\caption{Segmentation \& patch extraction of the CS-Image}
	\label{fig:segmentation}
\end{figure}
 \begin{figure*}[t!]
 	\centering
 	\subfigure[HLDB$_{SM}$]{\includegraphics[width=0.15\textwidth]{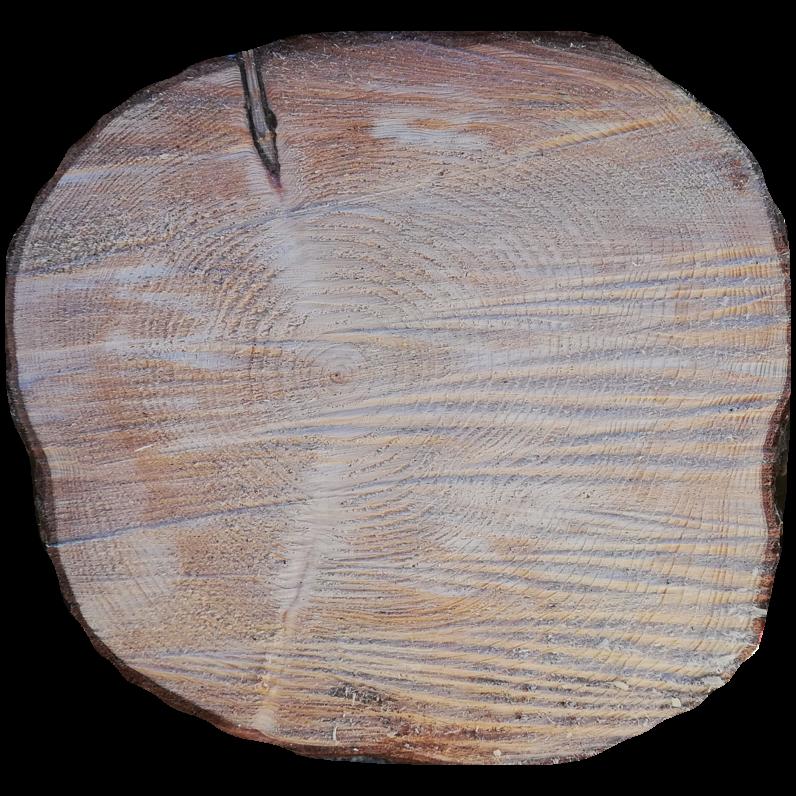}}
 	\subfigure[HLDB$_{SM}$]{\includegraphics[width=0.15\textwidth]{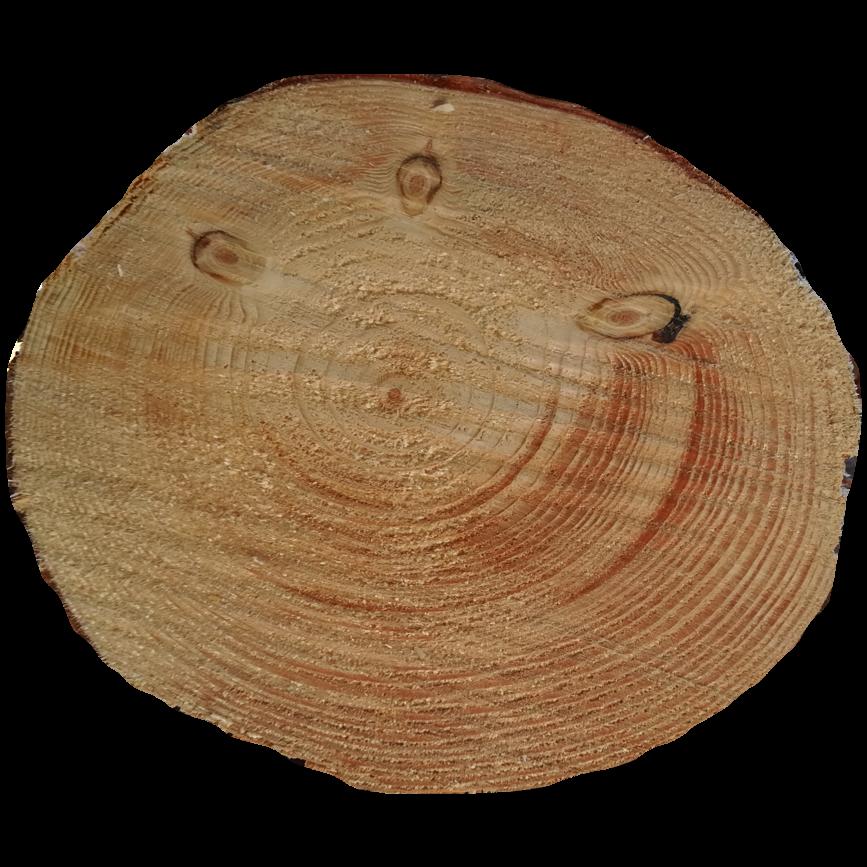}}	
 	\subfigure[HLDB$_{FH}$]{\includegraphics[width=0.15\textwidth]{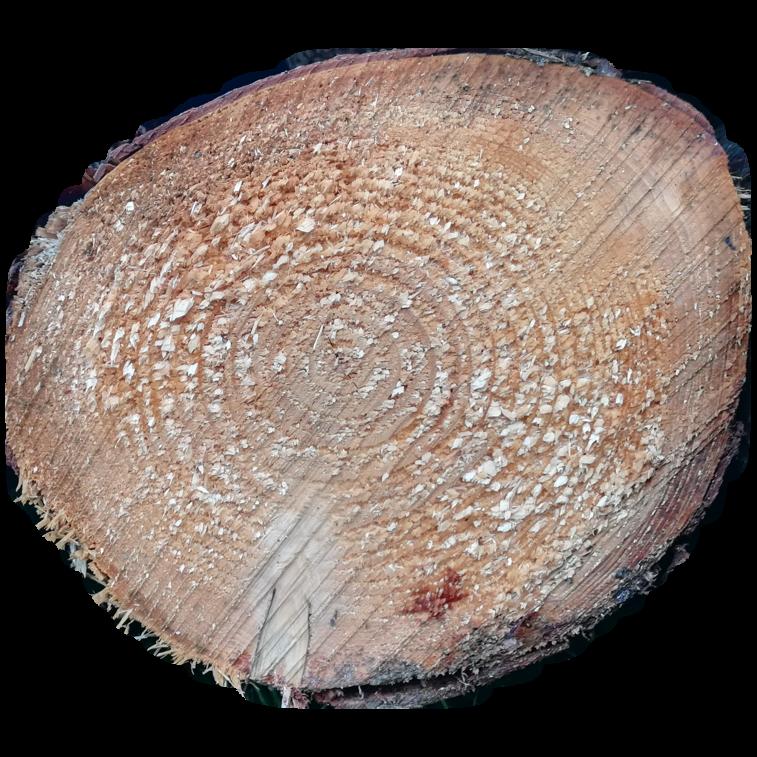}}
 	\subfigure[HLDB$_{FH}$]{\includegraphics[width=0.15\textwidth]{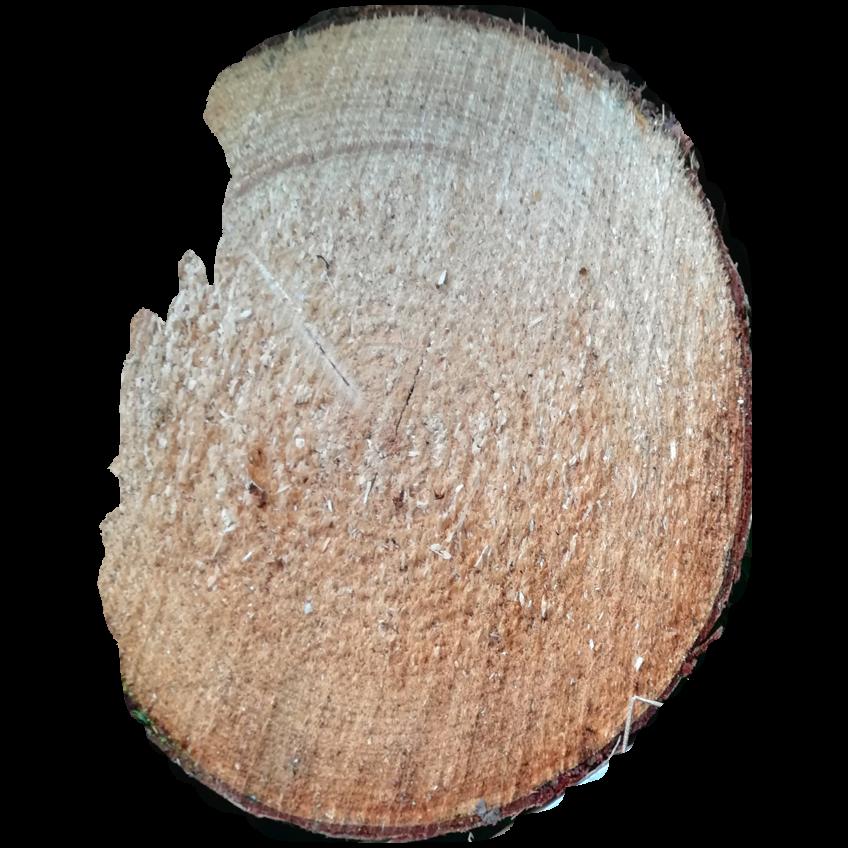}}
 	\subfigure[HLDB$_{FL}$]{\includegraphics[width=0.15\textwidth]{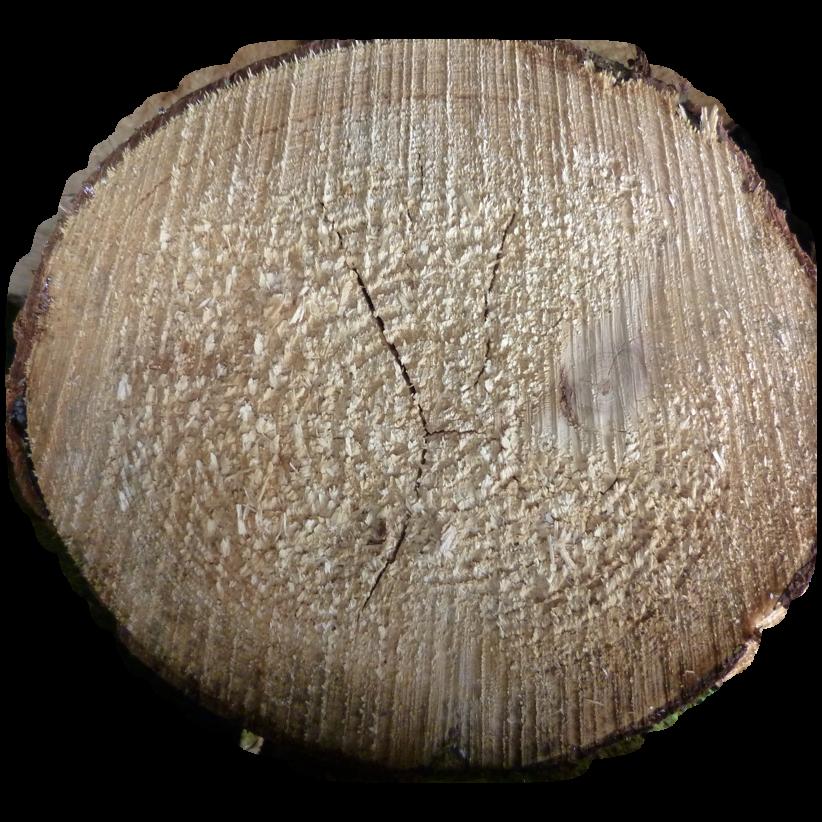}}
 	\subfigure[HLDB$_{FL}$]{\includegraphics[width=0.15\textwidth]{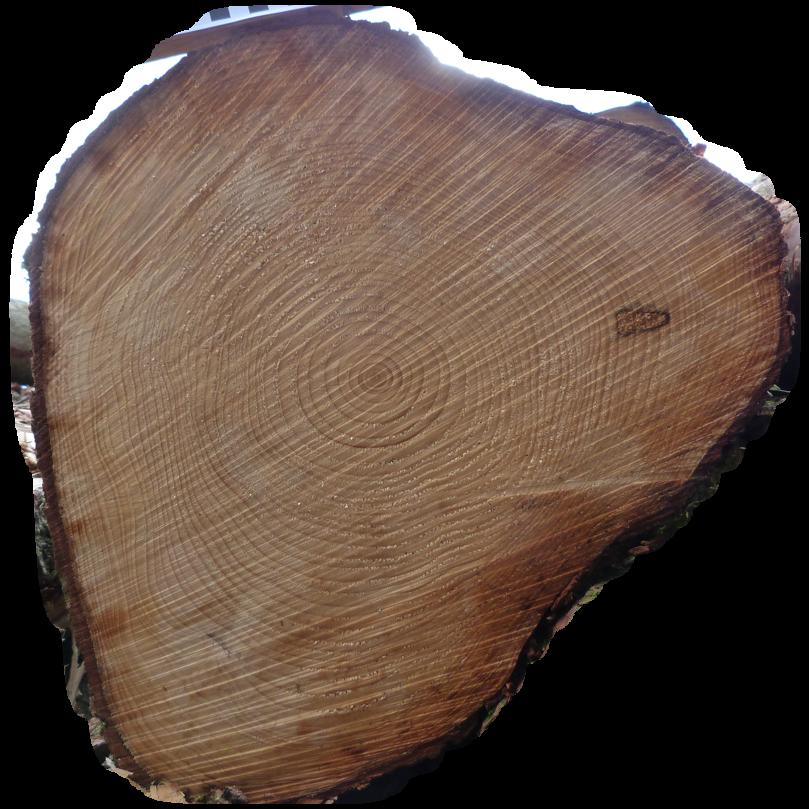}}
 	\caption{Exemplary outcomes of the segmentation in combination with square image patch extraction}
 	\label{fig:degresult}
\end{figure*}
%\begin{figure}[t!]
%	\centering
%	\subfigure[HLDB$_{SM}$]{\includegraphics[width=0.15\columnwidth]{fva_huawei_E006B_1}}
%	\subfigure[HLDB$_{SM}$]{\includegraphics[width=0.15\columnwidth]{fva_huawei_E007H_1.jpg}}	
%	\subfigure[HLDB$_{FH}$]{\includegraphics[width=0.15\columnwidth]{forest_huawei_E042H_1}}
%	\subfigure[HLDB$_{FH}$]{\includegraphics[width=0.15\columnwidth]{forest_huawei_E040H_3}}
%	\subfigure[HLDB$_{FL}$]{\includegraphics[width=0.15\columnwidth]{forest_lumix_E017H_2}}
%	\subfigure[HLDB$_{FL}$]{\includegraphics[width=0.15\columnwidth]{forest_lumix_E012B_3}}
%	\vspace{-3mm}
%	\caption{Exemplary outcomes of the segmentation in combination with square image patch extraction}
%	\label{fig:degresult}
%\end{figure}
The obtained segmentation mask of a CS-Image, which consists of probability values between 0 and 1 for each pixel of the image, is binarized. All values of the CNN segmentation mask that are below the threshold value $t$ ($t=0.5$ for HLDB$_{SM}$ and MVA and $t=0.25$ for HLDB$_{FH,FL}$) are set to zero and the remaining values are set to one.
The binarized segmentation mask of the CS is further used to set the background (all image positions with a zero in the segmentation mask) to black. Finally, each CS-Image is reduced to the smallest possible square shaped image section so that the CS (all image positions with a '1' in the segmentation mask) is still completely included in the image together with a five pixel thick black border on each side of the image. A schematic representation of the segmentation including the extraction of the square shaped image patch containing the CS is displayed in Fig.~\ref{fig:segmentation}. 
For MVA we can quantitatively assess the outcome of the segmentation. Averaged over all CS-Images in MVA, $99.26\%$ of the pixels per image were correctly segmented. In Figure \ref{fig:degresult}, we present exemplar outcomes of the segmentation and patch extraction process for the two forest DBs and the HLDB$_{SM}$.
The segmentation outcomes on the HLDB$_{SM}$ all look perfectly fine based on the authors' visual impression. For the two forest datasets HLDB$_{FH,FL}$, most images were well segmented, but on some images, parts of the log CS were predicted as background which was the reason why we used a smaller threshold ($t=0.25$ instead of 0.5) to binarize the segmentation masks. This reduced the risk to predict parts of the CS as background (as can be seen in Figure \ref{fig:degresult}(d)) but also 
led to the problem that for some images a bit of background surrounding the log CS was predicted as being part of the log CS (see Fig.~\ref{fig:degresult}(f)).

The advantage of our proposed segmentation and square image patch extraction approach for log recognition is that the background of a log CS image does not influence the log recognition. For CNN based recognition systems, where the images usually have to be resized to a fixed size before feeding them through the network, an additional advantage is the reduced loss of image quality. % by resizing the input image to the required CNN input size.
The segmented square shaped image patches are clearly smaller than the original CS-Image and so less information on the log is lost by reducing the image resolution to fit the required CNN input size.

\section{Wood log recognition using CNN triplet loss}
\label{sec:recognition}

In biometric applications, the problem with common CNN loss functions (e.g. the SoftMax loss) is that CNNs are only able to identify those subjects which have been used for the training of the neural network. If new subjects are added in a biometric application system, then the nets need to be trained again or else a new subject can only be classified as one of the subjects that were used for training (the one that is most similar to the newly added subject with respect to the CNN). This of course renders the practical application of common CNN loss functions impossible for biometric applications.

Contrary to more common loss functions like the Soft-Max loss, the triplet loss \cite{Schroff15} does not directly learn the CNN to classify images to their respective classes. %It does not even require to know the class affiliation of the employed training images.
The triplet loss requires three input images at once (a so called triplet), where two images belong to the same class (the so called Anchor image and a sample image from the same class, further denoted as Positive) and the third image belongs to a different class (further denoted as Negative). The triplet loss learns the network to minimize the distance between the Anchor and the Positive and maximize the distance between the Anchor and the Negative. 
% The triplet loss using the squared Euclidean distance is defined as follows:
% \begin{equation}
% 	\begin{aligned}
% 		L(A,P,N)=	& \max(||f(A)-f(P)||^{2} \\
% 					& -||f(A)-f(N)||^{2} +\alpha, 0),
% 	\end{aligned}
% 	\label{eq:defTripletLoss}
% \end{equation}
% where $A$ is the Anchor, $P$ the Positive and $N$ the Negative. $\alpha$ is a margin that is enforced between positive and negative pairs and is set to $\alpha=1$. $f(x)$ is an embedding (the CNN output) of an input image $x$.
% \begin{figure}[bt]
% 	\centering
% 	\includegraphics[width=0.98\columnwidth]{TripletLoss}
% 	\caption{CNN training using the triplet loss}
% 	\label{fig:CNNTriplet}
% \end{figure}
% Figure~\ref{fig:CNNTriplet}
% shows the scheme of learning a CNN using the triplet loss. 
% A triplet of training images (Anchor, Positive and Negative) is fed through the CNN resulting in an embedding for each of the three images.
% The embeddings of the three images are then used to compute the triplet loss to update the CNN.
\begin{table*}[!t]
	\centering
	\footnotesize{
		\begin{tabular}{p{0.17cm}|c|c|c|c|c|c|c|c}
			& Methods&MVA& HLDB$_{FH}$	& HLDB$_{FL}$&HLDB$_{SM}$& HLDB$_{S}$ &HLDB$_{R}$\\ \hline
			\multirow{2}{2cm}{\rot{CNN}} &SqNet&0.7 / 1.0 &3.2&2.4&3.1&3.4& 2.8\\ 
			&SqNet+&0.6 /1.0&2.8 &1.7&2.6&3.4&2.6\\ \hline
			\multirow{2}{2cm}{\rot{Traditional}}&Iris$_H$& 2.12 \cite{Schraml16c} /8.7&34.0&34.0&24.9&8.5 &13.5\\
			&Iris$_V$& 0.9 \cite{Schraml16c} / 5.4&29.1 &28.9&21.8&5.1&11.4\\
			&FP$_{CG}$& - /3.9 & 16.9 & 19.5 & 20.3 & 8.0 & 8.3 \\
			%&FP$_{RGP}$& TD & TD & TD & TD & TD & TD \\
	\end{tabular} }
	\caption{Recognition performance (Mean EER in [\%]) of the methods on the 5 log CS datasets. %proposed method using the two different training strategies, as well as four approaches using classical hand-crafted biometric features, on the 5 log CS datasets. 
		On the MVA dataset, recognition is applied on the manually segmented images (value before the slash) and the CNN segmented images (value after slash).}
	\label{table:result}
	
\end{table*}

%Summarized this means the CNN is trained to create an embedding $f(x)$, from an image to the feature space $\mathbb R^{d}$, such that the squared distances between all log CS images of the same class (log) is small, whereas the squared distance between any pairs of log images from different logs is large. 
%As image representation (feature vector) for the classification of a log image we employ the embedding $f(x)$ of the log image $x$. 
For our application this means that the CNN is trained so that the Euclidean distances between the CNN feature vectors of all log CS-Images of the same class (log) is small, whereas the Euclidean distance between any pairs of CS-Images from different logs is large.
%Similar to a publication on finger vein recognition \cite{Wimmer20a}, we employ hard triplet selection (only those triplets are chosen for training that actively contribute to improving the model)
%and the Squeeze-Net (SqNet) architecture \cite{Moskewicz16}, a small neural networks that is specifically created to have few parameters and only small memory requirements.
We employ hard triplet selection \cite{Schroff15} (only those triplets are chosen for training that actively contribute to improving the model)
and the Squeeze-Net (SqNet) architecture \cite{Moskewicz16}. SqNet is a small neural networks that is specifically created to have few parameters and only small memory requirements.

The size of the CNN's last layer convolutional filter is adapted so that a 256-dimensional output vector (embedding) is produced.
To make the CNN more invariant to shifts and rotations and increase the amount of training data, we employ data augmentation for CNN training. The images are randomly rotated in the range of 0-360 \textdegree{} and random shifts in horizontal and vertical directions are applied by first resizing the input images to a size of $234\times234$ and then extracting a patch of size $224\times 224$ (the best working input size using the SqNet for log recognition) at a random position of the resized image ($\pm 10$ pixels in each direction).
The CNN is trained for 400 epochs, starting with a learning rate of 0.001 that is divided by 10 every 120 epochs.

\section{Experimental Setup}

\label{sec:setup}
In this work, a 4-fold cross validation is employed. For each dataset, the CNN is trained four times, each time using three of the folds for training and evaluation is applied on the remaining fold. Each fold consists of a fourth of the logs of a dataset, where all images of a log are in the same fold. We further denote these experiments as ``SqNet''.
In a second experiment, we additionally use training data from logs of other datasets. For the HLDB datasets, we additionally use MVA for training (they are added to the three training folds).
%for each of the four folds we use all the images of the MVA dataset additionally to the three training folds for training. %(The MVA dataset is too different to the CT dataset and so it would not work to additionally use MVA data for the training of the CT dataset.)
In case of MVA, we additionally employ all the images of HLDB$_{FH}$ for training. We further denote these experiments with additional training data from another dataset as ``SqNet+''. For performance evaluation we have decided to present verification results, i.e. we compute the Equal Error Rates (EERs) for the different datasets achieved with SqNet and SqNet+. 

%neu
%Although the actual application corresponds to the identification scenario, 

%The reason for this is that each dataset has a different number of images per log end which differ more or less from one another. The EER, which is based on the genuine and impostor comparison score distributions, considers all comparisons of CS-Images in a dataset and it enables to consider the principal distinctivity between different logs. 
The EER is well suited to compare the CNN-based results to results achieved with traditional approaches and with results achieved in prior works (e.g. in
\cite{Schraml16c}).
%the future. On the contrary, the computation of identification results (e.g. Rank-1 recognition rate) depends on the selection of probe and query templates for each log which is often done using different protocols, making it difficult to compare the results with results of others.
%Des müss ma noch rechtfertigen!!!
%As performance measures, we compute the equal error rate (EER), the error rate of a verification system when the operating threshold for the accept/reject decision is adjusted such that the probability of false acceptance and that of false rejection become equal. % and the accuracy using a 1-Nearest-Neighbor classifier.
%We employ the genuine and imposter scores of all genuine (image pairs from the same log and log side) and imposter comparisons (image pairs from different logs) for EER computation. 
%All possible genuine and imposter comparisons are performed to calculate the genuine and imposter scores for the EER computation. %For calculating the impostor scores, only the first image of a finger is compared against the first image of all other fingers.
We have to consider that each of the four trained CNNs per dataset (one per fold) has a different mapping of the images to the CNN output feature space. Thus, feature vectors of different folds cannot be compared in the evaluation and the EER %and accuracy have
has to be computed separately for each fold.
We report the mean EER 
%and the mean accuracy 
over the four folds. 

As already mentioned before, the HLDB datasets consist of images from both log ends which show no obvious visible similarities. To employ the maximum number of images for CNN training, both ends are considered as different classes thus resulting in 200 classes in total. To avoid any bias by assigning different classes to the two sides of a log, we exclude those triplets during training where the Anchor and the Negative are from the same log but different sides. The same is applied for EER computation, where those comparison scores (scores between images from different sides of the same log) are ignored. 
%between images from different sides of a log for EER computation %and also exclude the distances between images of different sides of the same log to determine the nearest neighbors. 
%(Also in practical application, no one would compare images of different sides of a log since images would only be taken from one side of a log but not from both).

%In addition to the experiments for each dataset separately, we also want to know if a cross-dataset recognition is possible for datasets showing the same log CS. Only the two the two Forest datasets show the same log CS, all other databases show different log CS because the ends of the logs were cut (HLDB$_{SM}$) or images of the log CS were gathered inside the log instead of at the ends of the log). For the cross dataset experiment, we combine the images of the two Forest datasets to one dataset and use the same experimental setup on the new combined databset as for the previous experiments. For the computation of the EER, we additionally perform a cross-dataset experiment where we only use those similarity scores that are between images of different datasets, so that an image of one DB can only be recognized by means of the images of the other dataset. 

\paragraph*{Comparison Methods:}
In order to assess the performance of CNN-based wood log recognition we compute EERs using the fingerprint- and iris-based approaches proposed in \cite{Schraml16c}. The iris-based results IRIS$_V$ and IRIS$_H$ are computed in the exact same way as described in \cite{Schraml16c}. 
For rotational pre-alignment the CM (center of mass to pith estimate vector) strategy is applied, features are computed with the Log Gabor configuration LG (64/08) and %for 
matching code shifting is done in the range of -21 to 21 feature vector positions. %Note, that for IRIS$_V$ shifting corresponds to a movement in radial direction. 
For the fingerprint-based approach we utilize a modified approach, based on a circular grid, as introduced in \cite{Schraml20a}, which does not require to compute feature vectors for rotated versions of the registrated CS-Image. 
%Identical as for the iris-based approach in the template comparison procedure rotation compensation is performed by shifting the feature vectors of each band. 
Identical as the template comparison procedure for the iris-based approach, rotation compensation is performed by shifting the feature vectors of each band.
%There is no space and need to describe the shifting procedure in detail. 
This circular grid fingerprint-based approach is referred to as FP$_{CG}$. Contrasting to our previous work, we do not utilize manually extracted groundtruth data and instead use the CNN-based CS-Segmentation results and the pith position determined using the approach described in \cite{Schraml15b}. 
%no groundtruth data for the CS boundary and pith position was utilized. For CS boundary estimation, the CNN-based CS-Segmentation results are utilized and the pith position is determined using the approach described in \cite{Schraml15b}. The mean EERs are computed using the same protocol as utilized for the CNN methods. %-based mean EERs.

\section{Results}
\label{sec:results}

In Table~\ref{table:result}, we present the EER rates of the CNN-based approaches and the traditional approaches for each dataset. The main finding in Table~\ref{table:result} is that the CNN-based approaches are clearly superior to the traditional ones. %approaches. 
SqNet+ performs slightly better than SqNet due to the higher amount of training data. 
Considering the MVA results, an EER of 0.9\% achieved by IRIS$_V$ was the best result presented in \cite{Schraml16c}. 
The results for the CNN-based approaches using segmentation groundtruth data (SqNet = 0.7\%, SqNet+ = 0.6\%) outperform our previous results and the EERs achieved with the automated CS-Segmentation (SqNet/+ = 1\%) are close to it.
%On the one hand, the results for the CNN-based approaches using segmentation groundtruth data (SqNet = 0.7\%, SqNet+ = 0.6\%) outperform our previous results and on the other hand the EERs achieved with the automated CS-Segmentation (SqNet/+ = 1\%) are close to it.
Thus, another main advantage of this work is that the proposed CS-Segmentation is well suited to be used with the CNN-based recognition approaches. This statement is confirmed by the EERs presented for the HLDB datasets, which are all below 3.4\%. The EERs presented for the traditional approaches are not even close to this performance. By comparing the EERs for the HLDB computed with the traditional approaches it is obvious that the EERs computed for HLDB$_{S,R}$ are better than those computed for HLDB$_{FL,FH,SM}$. The reason is that for HLDB$_{S,R}$ rotational pre-alignment is more accurate than for the other datasets
because of the more accurate CNN segmentation for HLDB$_{S,R}$. 
%This has two reasons: (i) The CNN-based segmentation for HLDB$_{S,R}$ and MVA is more accurate than for HLDB$_{FL,FH,SM}$ and 
%(ii) The first reason causes that the applied rotational pre-alignment strategy which depends on the pith position to CS center of mass vector gets inaccurate.
%(ii) The first reason causes that the applied rotational pre-alignment strategy, which depends on the position of the pith compared to the CS's center of mass, gets inaccurate.
However, these observations highlight the main advantage of the CNN-based approaches: They works in combination with a fully automated CNN-based segmentation and do not require
%to rotationally pre-align the CS prior 
any rotational pre-alignment prior to feature extraction. 
%In Table \ref{table:resultcross}, we present the results of our cross-database experiment using the two Forest databases.
%\begin{table}[htb]
%	\begin{center}		
%		\begin{tabular}{c c c}
%			Methods				 &Forest HLDB & Forest cross-database evaluation 		\\ \HLDBine
%			SqNet    & 1.8 &2.4  \\ 
%			SqNetTrain+   &1.9  &2.6
			
%		\end{tabular} 
%	\end{center}
%	\caption{Recognition performance (EER in [\%]) for the combination of the images of the two Forest DBs using either all similarity scores (Forest HLDB) or only the similarity scores between images of different databases (Forest cross-database evaluation)}
%	\label{table:resultcross}
%\end{table}

\section{Conclusion}

\label{sec:conclusions}
Recently, there has been an increasing interest in methods for tracking roundwood on the basis of each individual log. In prior works we proposed a physical free approach using log end images and methods inspired by fingerprint and iris recognition-based approaches. Results were promising and showed good performances when using groundtruth data for segmentation of the log end in each image. However, in a real world application a fully automated system is required. In order to close this gap, we employ a CNN-based segmentation approach combined with a CNN-based log recognition approach which is compared to results achieved with the traditional log recognition approaches when using automatically segmented images. Results showed, that the CNN-based wood log recognition works well in combination with the CNN-based segmentation. On the contrary, the traditional approaches suffer from the inaccuracies of the CNN-based segmentation which affects the required rotational pre-alignment strategy. It can be concluded that the proposed two-stage CNN-based wood log recognition approach is well suited for individual wood log tracking. What remains is to prove that this two-stage approach aslo works in a realistic scenario, i.e. if logs can be tracked when using imagery captured at various stages of the log tracking chain.
% IEEE

%\section{Acknowledgments}
%This work is partially funded by the Austrian Science Fund (FWF) under Project number I 3653.

% LNCS
%\bibliographystyle{splncs}

% IEEE
\bibliographystyle{IEEE.bst}
\bibliography{bibs2,all,wood}

\begin{thebibliography}{10}

\bibitem{Tzoulis2013}
Ioakeim Tzoulis and Zaharoula Andreopoulou,
\newblock ``Emerging traceability technologies as a tool for quality wood
  trade,''
\newblock {\em Procedia Technology}, vol. 8, no. 0, pp. 606--611, 2013.

\bibitem{Schraml15d}
Rudolf Schraml, Johann Charwat-Pessler, Alexander Petutschnigg, and Andreas
  Uhl,
\newblock ``Towards the applicability of biometric wood log traceability using
  digital log end images,''
\newblock {\em Computers and Electronics in Agriculture}, vol. 119, pp.
  112--122, 2015.

\bibitem{Schraml16a}
Rudolf Schraml, Johann Charwat-Pessler, Karl Entacher, Alexander Petutschnigg,
  and Andreas Uhl,
\newblock ``Roundwood tracking using log end biometrics,''
\newblock in {\em Proceedings of the Annual GIL Meeting (GIL'2016)}. 2016, LNI,
  pp. 189--192, Gesellschaft für Informatik.

\bibitem{Tang2018}
Xin~Jie Tang, Yong~Haur Tay, Nordahlia~Abdullah Siam, and Seng~Choon Lim,
\newblock ``{MyWood}-{ID},''
\newblock in {\em Proceedings of the 2018 International Conference on
  Computational Intelligence and Intelligent Systems - {CIIS} 2018}. 2018,
  {ACM} Press.

\bibitem{Olschofsky2020}
Konstantin Olschofsky and Michael K\"{o}hl,
\newblock ``Rapid field identification of cites timber species by deep
  learning,''
\newblock {\em Trees, Forests and People}, vol. 2, pp. 100016, dec 2020.

\bibitem{Fricker2019}
Geoffrey~A. Fricker, Jonathan~D. Ventura, Jeffrey~A. Wolf, Malcolm~P. North,
  Frank~W. Davis, and Janet Franklin,
\newblock ``A convolutional neural network classifier identifies tree species
  in mixed-conifer forest from hyperspectral imagery,''
\newblock {\em Remote Sensing}, vol. 11, no. 19, pp. 2326, oct 2019.

\bibitem{Hu2019}
Junfeng Hu, Wenlong Song, Wei Zhang, Yafeng Zhao, and Alper Yilmaz,
\newblock ``Deep learning for use in lumber classification tasks,''
\newblock {\em Wood Science and Technology}, vol. 53, no. 2, pp. 505--517, feb
  2019.

\bibitem{Schraml16c}
Rudolf Schraml, Heinz Hofbauer, Alexander Petutschnigg, and Andreas Uhl,
\newblock ``On rotational pre-alignment for tree log end identification using
  methods inspired by fingerprint and iris recognition,''
\newblock {\em Machine Vision and Applications}, vol. 27, no. 8, pp.
  1289--1298, 2016.

\bibitem{Schraml15c}
Rudolf Schraml, Heinz Hofbauer, Alexander Petutschnigg, and Andreas Uhl,
\newblock ``Tree log identification based on digital cross-section images of
  log ends using fingerprint and iris recognition methods,''
\newblock in {\em Proceedings of the 16th International Conference on Computer
  Analysis of Images and Patterns (CAIP'15)}, Valetta, MLT, 2015, LNCS, pp.
  752--765, Springer Verlag.

\bibitem{He2017}
Kaiming He, Georgia Gkioxari, Piotr Dollar, and Ross Girshick,
\newblock ``Mask r-{CNN},''
\newblock in {\em 2017 {IEEE} International Conference on Computer Vision
  ({ICCV})}. oct 2017, {IEEE}.

\bibitem{Schroff15}
F.~{Schroff}, D.~{Kalenichenko}, and J.~{Philbin},
\newblock ``Facenet: A unified embedding for face recognition and clustering,''
\newblock in {\em 2015 IEEE Conference on Computer Vision and Pattern
  Recognition (CVPR)}, June 2015, pp. 815--823.

\bibitem{Moskewicz16}
Forrest~N. Iandola, Matthew~W. Moskewicz, Khalid Ashraf, Song Han, William~J.
  Dally, and Kurt Keutzer,
\newblock ``Squeezenet: Alexnet-level accuracy with 50x fewer parameters and
  {\textless}1mb model size,''
\newblock {\em CoRR}, vol. abs/1602.07360, 2016.

\bibitem{Schraml20a}
Rudolf Schraml, Karl Entacher, Alexander Petutschnigg, Timothy Young, and
  Andreas Uhl,
\newblock ``Matching score models for hyperspectral range analysis to improve
  wood log traceability by fingerprint methods,''
\newblock {\em Mathematics}, vol. 8, no. 7, pp. 10, 2020.

\bibitem{Schraml15b}
Rudolf Schraml, Alexander Petutschnigg, and Andreas Uhl,
\newblock ``Validation and reliability of the discriminative power of geometric
  wood log end features,''
\newblock in {\em Proceedings of the IEEE International Conference on Image
  Processing (ICIP'15)}, Quebec, CAN, 2015.

\end{thebibliography}

\end{document}